\documentclass{article}
\usepackage{amssymb}
\usepackage[table,xcdraw]{xcolor} 

\usepackage[preprint]{corl_2025} % Uncomment for pre-prints (e.g., arxiv); This is like ``final'', but will remove the CORL footnote.
\usepackage{xspace}
\usepackage{graphicx}
\usepackage{bm}
\usepackage{multirow}
\usepackage{wrapfig}
\usepackage{subfig}
\usepackage{amsmath}
\usepackage{textcomp}
\usepackage{enumitem}

\usepackage{url}            % simple URL typesetting
\usepackage{booktabs}       % professional-quality tables
\usepackage{amsfonts}       % blackboard math symbols
\usepackage{nicefrac}       % compact symbols for 1/2, etc.
\usepackage{microtype}      % microtypography
\usepackage{makecell}
\usepackage[ruled,noend,linesnumbered]{algorithm2e}
\definecolor{darkgreen}{rgb}{0.09, 0.45, 0.27}
\definecolor{snsorange}{RGB}{255, 141, 98}
\definecolor{ourblue}{RGB}{216,237,243}
\definecolor{ourred}{RGB}{249,214,214}
\definecolor{ourred2}{RGB}{221,73,57}

\definecolor{tablecolor}{RGB}{255,224,179}
\definecolor{ourorange}{RGB}{255,153,0}
\definecolor{tablecolor2}{RGB}{240, 220, 220}
\newcommand{\dd}[2]{$#1\scriptstyle{\pm#2}$}

\newcommand{\singleddbf}[1]
{\cellcolor{ourblue}$\mathbf{#1}$}

\newcommand{\ddbf}[2]
{\cellcolor{ourblue}$\mathbf{#1\scriptstyle{\pm#2}}$}

\newcommand{\ccbf}[1]{\cellcolor{ourred}$\mathbf{#1}$}
\newcommand{\best}[1]{\textcolor{ourred2}{\singleddbf{#1}}}

\title{H$^{\mathbf{3}}$DP: Triply‑Hierarchical Diffusion Policy for Visuomotor Learning}
\newcommand{\ourshort}{H$^{3}$DP\xspace}
\newcommand{\bfourshort}{\textbf{H$^{\mathbf{3}}$DP}\xspace}

\author{
Yiyang Lu$^{1*}$, \quad
Yufeng Tian$^{4*}$, \quad
Zhecheng Yuan$^{1,2,3}\thanks{Equal Contribution}$, \quad
Xianbang Wang$^{1}$, \quad \\
\textbf{Pu Hua$^{1,2,3}$, \quad}
\textbf{Zhengrong Xue$^{1,2,3}$, \quad} 
\textbf{
Huazhe Xu$^{1,2,3}$} \\
$^{1}$ Tsinghua University IIIS, $^{2}$ Shanghai Qi Zhi Institute, \\ $^{3}$ Shanghai AI Lab, $^{4}$ Harbin Institute of Technology  \\
\texttt{luyy24@mails.tsinghua.edu.cn,
huazhe\_xu@mail.tsinghua.edu.cn}
}

\begin{document}
\maketitle

%===============================================================================
\begin{figure}[h]
  \centering
  % \vspace{-20pt}
  \includegraphics[width=1.0\linewidth]{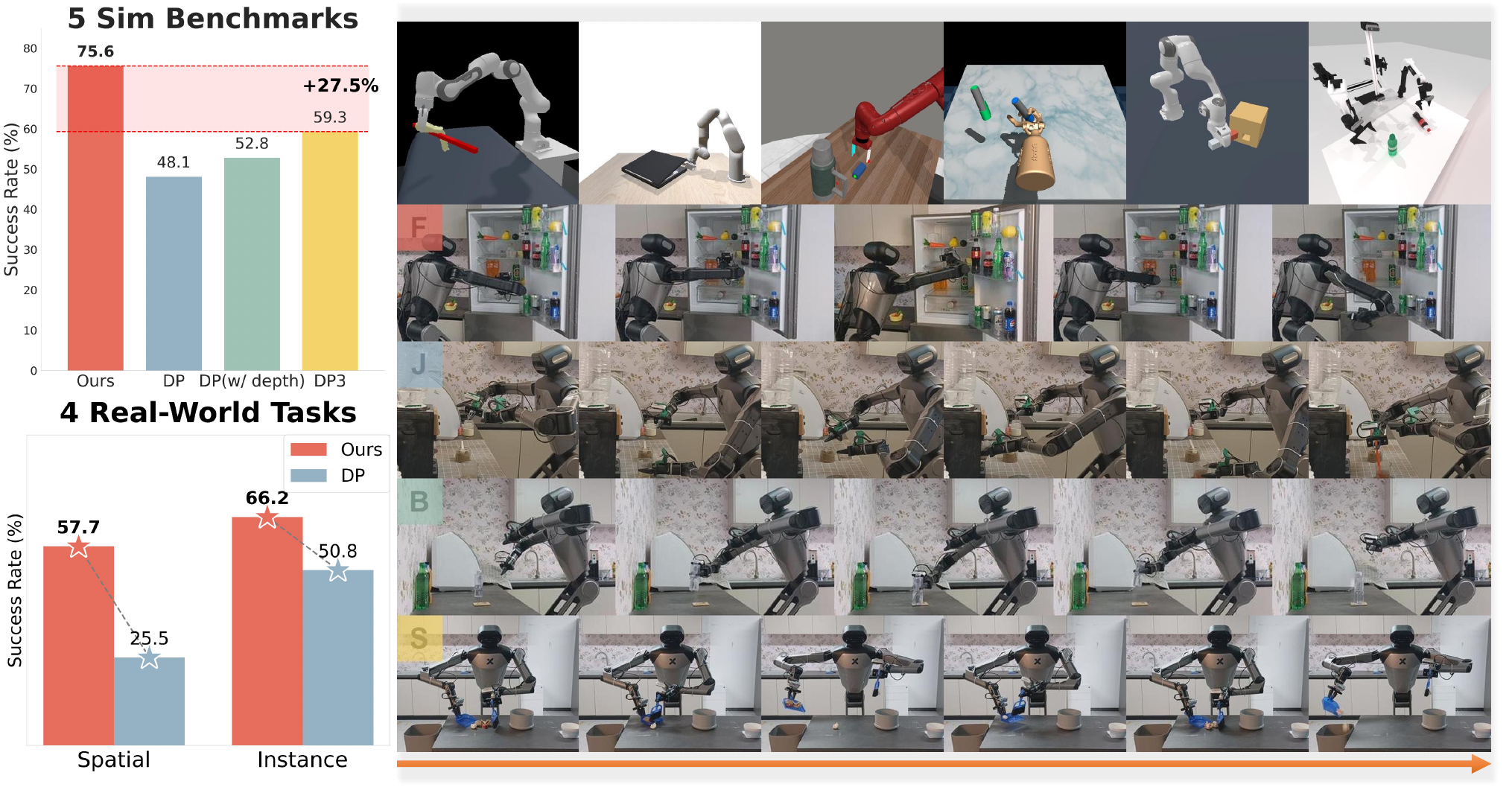}
  % \vspace{-15pt}
  \caption{\bfourshort can not only achieve superior performance across 44 tasks on 5 simulation benchmarks, but also handle long-horizon challenging manipulation tasks in cluttered real-world scenarios.}
  \label{fig:teaser}
\end{figure}

\begin{abstract}
Visuomotor policy learning has witnessed substantial progress in robotic manipulation, with recent approaches predominantly relying on generative models to model the action distribution. However, these methods often overlook the critical coupling between visual perception and action prediction. In this work, we introduce \textbf{Triply-Hierarchical Diffusion Policy}~(\bfourshort), a novel visuomotor learning framework that explicitly incorporates hierarchical structures to strengthen the integration between visual features and action generation. \ourshort contains $\mathbf{3}$ levels of hierarchy: (1) depth-aware input layering that organizes RGB-D observations based on depth information; (2) multi-scale visual representations that encode semantic features at varying levels of granularity; and (3) a hierarchically conditioned diffusion process that aligns the generation of coarse-to-fine actions with corresponding visual features. Extensive experiments demonstrate that \ourshort yields a $\mathbf{+27.5\%}$ average relative improvement over baselines across $\mathbf{44}$ simulation tasks and achieves superior performance in $\mathbf{4}$ challenging bimanual real-world manipulation tasks. Project Page: \url{https://lyy-iiis.github.io/h3dp/}.
\end{abstract}

\keywords{Imitation Learning, Representation Learning, Diffusion Model} 

%===============================================================================

\section{Introduction}
Visuomotor policy learning has emerged as a prevailing paradigm in robotic manipulation~\cite{chi2023diffusion, zhao2023learning, black2024pi0, Ze2024DP3, yuan2024learning}. Existing approaches have increasingly adopted powerful generative methods, such as diffusion and auto-regressive models, to model the action generation process~\cite{prasad2024consistency, wang2024one, frans2024one, shafiullah2022behavior, lee2024behavior}. However, these predominant methods have focused primarily on separately refining either the representation of perception or actions, often overlooking establishing a tight correspondence between perception and action. In contrast, human decision-making inherently involves hierarchical processing of information from perception to action~\cite{hubel1962receptive, bill2020hierarchical}. The visual cortex extracts features in a layered fashion and performs hierarchical inference based on visual motion perception, ultimately resulting in the generation of structured motor behaviors~\cite{lee2003hierarchical, bill2022visual}. Inspired by this, we argue that enabling learned visuomotor agents to emulate such hierarchical behavior patterns is also critical for enhancing their decision-making capabilities.

Prior works have primarily focused on hierarchically modeling the action generation process alone~\cite{su2025dense,gong2024carp}, without explicitly incorporating hierarchical structure throughout the whole visuomotor policy pipeline. In this paper, we present \bfourshort, a novel visuomotor policy learning framework grounded in three levels of hierarchy: input, representation, and action generation. This design reflects the hierarchical processing mechanisms that humans use the visual cortex to perceive environmental stimuli to guide motor behavior.

At the input level, to better leverage the depth information in modern robotic benchmarks and datasets~\cite{james2019rlbench, liu2023libero, sferrazza2024humanoidbench, geng2025roboverse}, \ourshort moves beyond prior 2D approaches that primarily rely on RGB or simple RGB-D concatenation, which has shown limited effectiveness in prior work~\cite{Ze2024DP3, zhu2024point}. We introduce \textbf{depth-aware layering} strategy that partitions the RGB-D input into distinct layers based on depth cues. This approach not only enables the policy to explicitly distinguish between foreground and background, but also suppresses distractors and occlusions~\cite{rao2010grasping,ainetter2021depth}, thereby enhancing the understanding and reasoning of spatial structure in the cluttered visual scenarios.

For visual representation, to address the limitations of flattening image features into a single vector, which can discard some spatial structures and semantic information~\cite{he2015spatial,ronneberger2015u,lin2017feature}, \ourshort employs \textbf{multi-scale visual representation}, where different scales capture features at varying granularity levels, ranging from global context to fine visual details. 

In the action generation stage, \ourshort incorporates a key inductive bias inherent to the diffusion process: the tendency to progressively reconstruct features from low-frequency to high-frequency components~\cite{rissanen2022generative,dieleman2024spectral,wang2025ddt}, by \textbf{hierarchical action generation}. Specifically, coarse visual features guide initial denoising steps to shape the global structure (low-frequency components) of action, while fine-grained visual features inform the later steps to refine precise details (high-frequency components). This establishes a tighter coupling between action generation and visual encoding, enabling the policy to generate actions that are semantically grounded in multi-scale perceptual features.

We validate \ourshort through extensive experiments on $\mathbf{44}$ simulation tasks across $\mathbf{5}$ diverse benchmarks, where it surpasses state-of-the-art methods by a relative average margin of $\mathbf{+27.5}\%$. Furthermore, in real-world evaluations, we deploy bimanual robotic systems to tackle four challenging tasks situated in cluttered environments, involving high disturbances and long-horizon objectives. \ourshort achieves a $\mathbf{+32.3}\%$ performance improvement over Diffusion Policy in these real-world scenarios.

\section{Related Work}

\textbf{Visual imitation learning.} Numerous studies have proposed efficient policy learning algorithms from different aspects~\cite{chi2023diffusion,zhao2023learning,xue2025demogen}. As a representative approach, to endow the learned policy multi-modality ability, Diffusion Policy~\cite{chi2023diffusion} incorporates the diffusion process to better represent the action distribution. Based on DP, methods like DP3~\cite{Ze2024DP3, ze2024humanoid_manipulation} and 3D-Actor~\cite{ke20243d}, designed for point cloud inputs, enhance the policy’s scene understanding by refining the visual representation. Consistency Policy~\cite{prasad2024consistency} and ManiCM~\cite{lu2024manicm} modify the inference process to achieve the inference acceleration. However, these approaches focus solely on enhancing either the action generation or the visual feature extraction, without explicitly modeling the relationship between them. To address this issue, we propose a hierarchical framework that couples multi-scale visual representations with the diffusion process, enabling a more structured integration between visual features and action generation.

\textbf{Leveraging hierarchical information for policy learning.} In the computer vision community, numerous studies have leveraged hierarchical information to address a variety of downstream tasks~\cite{van2017neural,wang2017multimodal,li2019hierarchical,razavi2019generating,liu2021swin,ryali2023hiera}. For example, standard diffusion models~\cite{song2019generative, ho2020denoising, song2020denoising, song2020score} and flow matching~\cite{lipman2022flow,liu2022flow,esser2024scaling} adopt the U-Net framework~\cite{ronneberger2015u,zhou2018unet++}, which exploits multi-scale feature representations to retain rich contextual information throughout the denoising process. VAR~\cite{tian2024visual} innovatively employs multi-scale visual representations with quantization to perform image generation in an auto-regressive manner. In robot learning, recent works~\cite{gong2024carp, pateria2021hierarchical, zhang2025autoregressive} have also begun to adopt hierarchical paradigms for policy learning. Dense Policy~\cite{su2025dense} leverages a bidirectional extension strategy to enable hierarchical action prediction. ARP~\cite{zhang2025autoregressive} predicts a sequence of actions at different levels of abstraction in a hierarchical way. CARP~\cite{gong2024carp} draws inspiration from VAR by employing a multi-scale VQ-VAE~\cite{van2017neural,razavi2019generating} to construct action sequences and subsequently generating residual actions autoregressively using a GPT-style architecture~\cite{radford2018improving}. However, these algorithms model only the hierarchical structure of the action generation process, without explicitly addressing the crucial linkage between visual representation and action in visuomotor policy learning. In contrast, \ourshort not only incorporates multi-scale visual representations but also leverages the inherent strengths of diffusion models to seamlessly integrate coarse-to-fine action generation into the diffusion process itself. Furthermore, by adopting a depth-aware layering strategy, \ourshort maximizes the utilization of hierarchical feature information across the input, latent, and output stages, thereby enriching the policy learning pipeline in a structured and semantically aligned manner.
\section{Method}

We employ three hierarchical structures to enhance the policy's understanding of visual input and predict more accurate action distributions. At the input level, the RGB-D image is discretized into multiple layers to improve the policy's ability to distinguish and interpret foreground-background variations. Upon this, we adopt a multi-scale visual representation, wherein coarse-grained features capture global contextual information, while fine-grained features encode detailed scene attributes. On the action side, correspondingly, the representations at different scales are utilized to generate actions in a coarse-to-fine manner, thus strengthening the correlation between action and visual representations. A detailed discussion of each part will be provided in the following sections.

\subsection{Depth-Aware Layering} 
\label{sec:depthlayer}

Effective robotic manipulation hinges on robust spatial understanding. While RGB data provides rich texture and color information, depth supplies the critical geometric context, including the relative spatial arrangement of objects and their distances. Combining these modalities offers a powerful foundation for scene comprehension. However, simply concatenating RGB images with depth maps does not lead to performance improvements~\cite{Ze2024DP3, zhu2024point}. Hence, to fully exploit the geometric structure inherent in depth maps, we introduce a depth-aware layering mechanism inspired by Zhang et al.~\cite{zhang2023monodetr}. Pixels with depth $d$ are assigned to layer $m$ using linear-increasing discretization:
\begin{equation}
\label{eq:layer}
m=\left\lfloor-0.5+0.5\sqrt{1+4(N+1)(N+2) \frac{d-d_{\min }}{d_{\max }-d_{\min }+\epsilon}}\right\rfloor,
\end{equation}
which promotes the robot to focus more on its workspace. By explicitly encoding objects distributed across different depth planes, this structured representation retains all visual detail while strategically utilizing depth to impose a meaningful foreground-background separation, thereby enabling the policy to selectively attend to different regions of the image. This design can effectively boost the agent's capacity for spatial perception and interaction planning. Furthermore, we also conduct comparisons against other discretization algorithms and perform additional experiments to substantiate the effectiveness of our proposed depth-aware layering method. The corresponding results are provided in Appendix~\ref{app:gmm} and Appendix~\ref{app:sig_depth}.

\begin{figure}[t]
  \centering
  \includegraphics[width=1.0\linewidth]{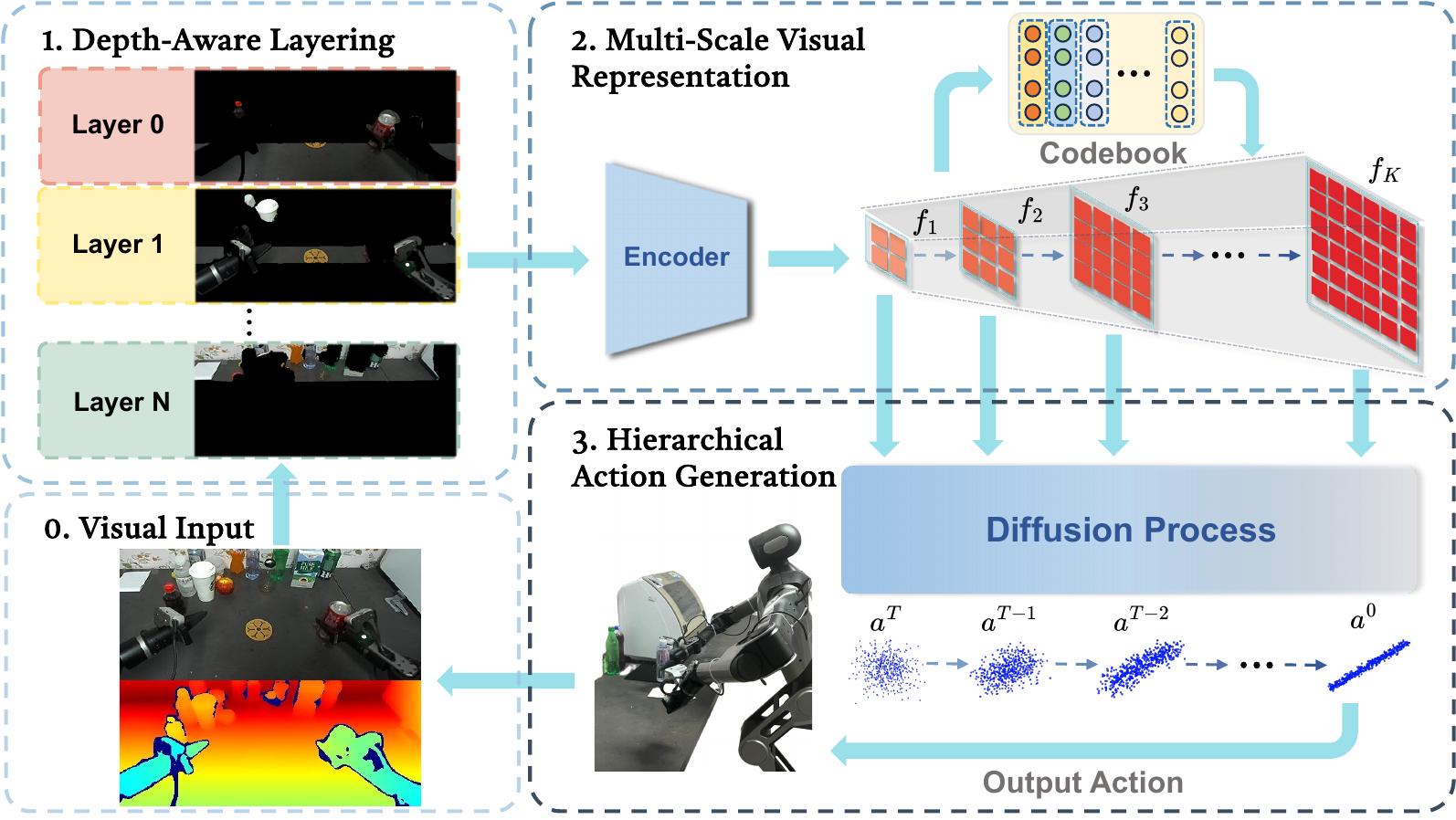}
  \caption{\textbf{Overview of \bfourshort}. \ourshort integrates three hierarchical design principles across the perception and action generation pipeline. At the input level, RGB-D images are decomposed into multiple layers based on their depth values. Then, we employ multi-scale visual representations to capture features at varying levels of granularity. During the action generation, denoising process is divided into several stages guided by multi-scale visual representations.}
  \label{fig:method}
\end{figure}

\subsection{Multi-Scale Visual Representation} 

In visuomotor policy learning, visual representation plays a crucial role in embedding input images and mapping them to actions. An effective visual encoder should capture various granularity features of the visual scenarios and guide the policy to predict the action distribution. However, existing methods typically extract features at a single spatial scale or compress them into a fixed-resolution representation, limiting the expressiveness of learned features~\cite{he2015spatial,ronneberger2015u,lin2017feature}. To address this problem, we hierarchically partition the feature map into multiple scales, enabling the capture of both coarse structural information and detailed local cues. 

\textbf{Interpolation and Quantization.} 
After applying depth-aware layering to the input image $I$, each layer $I_m$ is independently encoded into multi-scale feature maps $\{f_{m,k}|f_{m,k} \in \mathbb{R}^{h_{k} \times w_{k} \times C}\}_{k=1}^{K}$, where $\{(h_k,w_k)\}_{k=1}^K$ denotes the spatial resolutions across scales. Adopting the quantization design in VQ-VAE~\cite{van2017neural,razavi2019generating}, these feature maps $\{f_{m,k}\}_{k=1}^{K}$ are quantized into discrete vectors drawn from a learnable codebook $\mathcal{Z}_m \in \mathbb{R}^{V \times C}$. Specifically, each feature vector $f_{m,k}^{(i,j)}$ is mapped to its nearest neighbor in Euclidean distance:
\begin{equation}    
\label{eq:quant}
{f}^{(i,j)}_{m,k}\leftarrow \underset{z\in\mathcal{Z}_m}{\arg \min } \|z-f^{(i,j)}_{m,k}\|_2.
\end{equation}
By applying differentiable interpolation and lightweight convolution to the quantized features ${f}_{m,k}$, we then obtain the multi-scale visual representations $\{\hat{f}_{m,k}\}_{k=1}^K$ for each layer $I_m$. The pseudocode of full encoding procedure is detailed in Algorithm~\ref{alg:enc}, Appendix~\ref{app:method_details}.

\textbf{Training.} 
To ensure consistent representations across scales, we aim to minimize the consistency loss between the original feature $f_m=\mathcal{E}_m(I_m)$ and the representation $\hat{f}_{m,k}$ at different scales:
\begin{equation}
\label{eq:consistency}
    \mathcal{L}_{\text {consistency}}=\sum_{m=0}^{N-1}\sum_{k=1}^{K}\left(\left\|\hat{f}_{m,k}-\operatorname{sg}(f_m)\right\|_{2}^{2}+\beta\left\|f_m-\operatorname{sg}(\hat{f}_{m,k})\right\|_{2}^{2}\right),
\end{equation} 
where $\operatorname{sg}(\cdot)$ is the stop gradient operator and $\beta$ balances the gradient flow between two terms. The visual encoder $\{\mathcal{E}_m\}_{m=0}^{N-1}$ and codebook $\{\mathcal{Z}_m\}_{m=0}^{N-1}$ are trained end-to-end, as described in detail in Appendix~\ref{app:method_details}.

\subsection{Hierarchical Action Generation}

To match the inherent inductive biases of denoising process~\cite{rissanen2022generative,dieleman2024spectral,wang2025ddt}, we leverage multi-scale visual representations to model action generation in a coarse-to-fine manner. The early stage actions are derived from representations that capture global scene information, while fine-grained representations are responsible for generating detailed action components. This approach couples the visual representation and the action generation process via reinforcing their correspondence at the same hierarchical levels. 

\textbf{Inference.} 
Our action generation module is a denoising diffusion model conditioned on multi-scale features $F=\{\hat{f_k}=\{\hat{f}_{m,k}\}_{m=0}^{N-1}\}_{k=1}^{K}$ and robot poses $q$. The denoising process unfolds over $T$ steps partitioned into $K$ stages $\cup_{k=1}^{K}(\tau_{k-1},\tau_{k}]$. When $t\in (\tau_{k-1},\tau_{k}]$, the denoising network $\epsilon_\theta^{(t)}$ conditioning on the corresponding feature map $\hat{f}_k$ and robot poses $q$, predicts the noise component 
\begin{equation}
    \epsilon^t = \epsilon_{\theta}^{(t)}(a^t|\hat{f}_k, q),
\end{equation}
then generates $a^{t-1}$ from $a^t$ via:
\begin{equation}
    a^{t-1}=\sqrt{\alpha_{t-1}}\left(\frac{a^t-\sqrt{1-\alpha_t}\cdot{\epsilon}^t}{\sqrt{\alpha_t}}\right)+\sqrt{1 - \alpha_{t-1} - \sigma_t^2}\cdot{\epsilon}^t +\sigma_t \tilde{\epsilon}^t,
\end{equation}
gradually transforming the Gaussian noise $a^T$ into the noise-free action $a^0$, where $\alpha_t$, $\sigma_t$ are fixed parameters depending on the noise scheduler, and $\tilde{\epsilon}^t\sim\mathcal{N}(0,\mathbf{I})$ is a Gaussian noise. 

\textbf{Training.} 
To train the denoising network $\epsilon_\theta^{(t)}$, we randomly sample an observation-action pair $((I,q),a^0)\in \mathcal{D}$ and noise $\epsilon\sim\mathcal{N}(0,\mathbf I)$. The network is optimized to predict $\epsilon$ given a noisy action conditioned on the final feature map $\hat{f}_K$ and robot pose $q$, via the objective:
\begin{equation}
\label{eq:diffloss}
\mathcal{L}_{\text{diffusion}}=\mathbb{E}_{a^0,\epsilon,t}\left[\gamma_t\|\epsilon_\theta^{(t)}(\sqrt{\alpha}_t a^0+\sqrt{1-\alpha_t}\epsilon|\hat{f}_K,q)-\epsilon\|^2\right],
\end{equation}
where $\{\gamma_t\}$ are pre-defined coefficients. More implementation details can be found in Appendix~\ref{app:hyper}. By conditioning on the final feature $\hat{f}_K$ during training, gradients from the loss propagate through the entire hierarchical encoder, implicitly optimizing all $\{\hat{f}_k\}_{k=1}^K$. This design promotes consistency of representations at each scale for action generation while enhancing training efficiency.

\textbf{Discussions.} 
Diffusion models inherently aim to predict the posterior average of the target distribution conditioned on the provided features \cite{dhariwal2021diffusion,sun2025noise}, i.e., the optimal denoising network $\epsilon_{\theta^*}^{(t)}$ follows $\epsilon_{\theta^*}^{(t)}(a^t|f,q)=\mathbb{E}_{t,\epsilon,a^0,\sqrt{\alpha}_t a^0+\sqrt{1-\alpha_t}\epsilon=a^t}\left[\epsilon|a^t,f,q\right]$. Features at varying resolutions retain information across distinct frequency domains. Consequently, they provide robust guidance for generating specific frequency components of the action during relevant stages of the denoising process. Related experiments are shown in Section~\ref{sec:spectral}. By using lower-resolution features for earlier stages and gradually refining the predictions with higher-resolution features, the model benefits from both the stability of coarse representations and the precision of fine details. 
\section{Experiments}

In this section, we present extensive experiments across simulated and real‑world settings to demonstrate the efficacy of \ourshort.  In addition, we perform thorough ablation analyses to evaluate the contribution of each hierarchical design. 

\subsection{Simulation Experiments}

\subsubsection{Experiment setup}

\textbf{Simulation benchmarks and baselines:} 
To sufficiently verify the effectiveness of \ourshort, we evaluate \ourshort on \textbf{5} simulation benchmarks, encompassing a total of \textbf{44} tasks. These tasks span a variety of manipulation challenges, including articulated object manipulation~\cite{bao2023dexart,rajeswaran2017learning,yu2020meta}, deformable object manipulation~\cite{gu2023maniskill2}, bimanual manipulation~\cite{mu2024robotwin}, and dexterous manipulation~\cite{bao2023dexart,rajeswaran2017learning}. The details of the expert demonstrations can be found in Appendix \ref{app:expert_demo}. To comprehensively assess the performance of \ourshort, we compare it against three baselines: \textit{Diffusion Policy}~\cite{chi2023diffusion}, one of the most widely used visuomotor policy learning algorithms; \textit{Diffusion Policy~(w/ depth)}, which extends Diffusion Policy to incorporate RGB-D input to bridge the information gap; and \textit{DP3}~\cite{Ze2024DP3}, an enhanced version of Diffusion Policy that leverages an efficient encoder for point cloud input. 

\textbf{Evaluation metric:}
Each experiment is run with three different seeds to mitigate performance variance. For each seed, we evaluate 20 episodes every 200 training epoches. In simpler MetaWorld, Adroit and DexArt tasks, we compute the average of the highest five success rates as its success rate, while in other environments, only the hightest success rate is recorded. 

\begin{table}[t] \caption{\textbf{Simulation task results.} Across $5$ simulation benchmarks with various difficult levels, \ourshort obtains $\mathbf{+27.5\%}$ relative performance gains on average over 44 tasks. }
\label{table1: generalization}
\centering
\renewcommand\tabcolsep{4.0pt}
\begin{footnotesize}
\resizebox{1.0\textwidth}{!}{
\begin{tabular}{c|cccccccc|c}
\cline{2-6}
\toprule[0.5mm]
\begin{tabular}[c]{@{}c@{}}  \multicolumn{1}{c}{\multirow{2}{*}{Method  $\backslash$ Tasks}} \end{tabular} & MetaWorld & MetaWorld & MetaWorld & ManiSkill & ManiSkill & Adroit & DexArt & RoboTwin & \textbf{Average} \\
& \scriptsize (Medium 11) & \scriptsize (Hard 5) & \scriptsize (Hard++ 5) & \scriptsize (Deformable 4) & \scriptsize (Rigid 4) & \scriptsize (3) & \scriptsize (4) & \scriptsize (8) & \scriptsize $\mathbf{(44)}$ \\
\hline 
\begin{tabular}[c]{@{}c@{}} \bfourshort  \end{tabular} & \best{98.3}  & \best{87.8} & \best{95.8}  & \best{59.3} & \best{65.3} & \best{87.3} & 53.3  & \best{57.4} & \textcolor{ourred2}{\ddbf{75.6}{18.6}} \\
\begin{tabular}[c]{@{}c@{}} DP \end{tabular}  & $78.2$ & $52.6$  &  $58.0$  & $22.3$   &   $27.5$    & $79.0$   & $44.3$ &  $22.8$ &   \dd{48.1}{23.1}   \\
\begin{tabular}[c]{@{}c@{}}DP~(w/ depth)\end{tabular}& $77.7$   & $57.2$ & $71.2$ & $44.5$  &   $40.8$     &   $76.0$  &  $42.0$  & $12.6$ &  \dd{52.8}{22.2}  \\
\begin{tabular}[c]{@{}c@{}} DP3 \end{tabular} &  $89.1$ & $52.6$ & $88.4$ & $26.5$  &  $33.5$  & $84.0$ & \best{54.8} &  $45.9$  &  \dd{59.3}{24.9}  \\
\bottomrule[0.5mm]
\end{tabular}
}

\end{footnotesize}
\label{table:sim-results}
% \vspace{-10pt}
\end{table}

\subsubsection{Simulation performance}
\label{sec:sim_per}

As shown in Table~\ref{table:sim-results}, the simulation experiment results exhibit that \ourshort outperforms or achieves comparable performance among the whole simulation benchmarks. Our method outperforms DP3 by a relative average margin of $\mathbf{+27.5\%}$. Notably, DP3 requires manual segmentation of the point cloud to remove background and task-irrelevant elements. This process introduces additional human effort and renders performance susceptible to segmentation quality. Relevant experimental results are provided in Appendix~\ref{app:segdp3}. In contrast, benefiting from our design, \ourshort obtains superior performance using only raw RGB-D input, without the need for segmentation and human effort. Furthermore, on the Adroit and DexArt benchmark, while DP3 leverages multi-view cameras to restore the complete point clouds, \ourshort attains comparable performance using only one single-camera RGB-D image. The whole simulation results in each task can be found in Appendix~\ref{app:sim_total}.

\subsubsection{Spectral analysis of actions}
\label{sec:spectral}

\begin{wrapfigure}[13]{r}{0.63\textwidth}%
    \centering
    \vspace{-40pt}
    \includegraphics[width=0.63\textwidth]{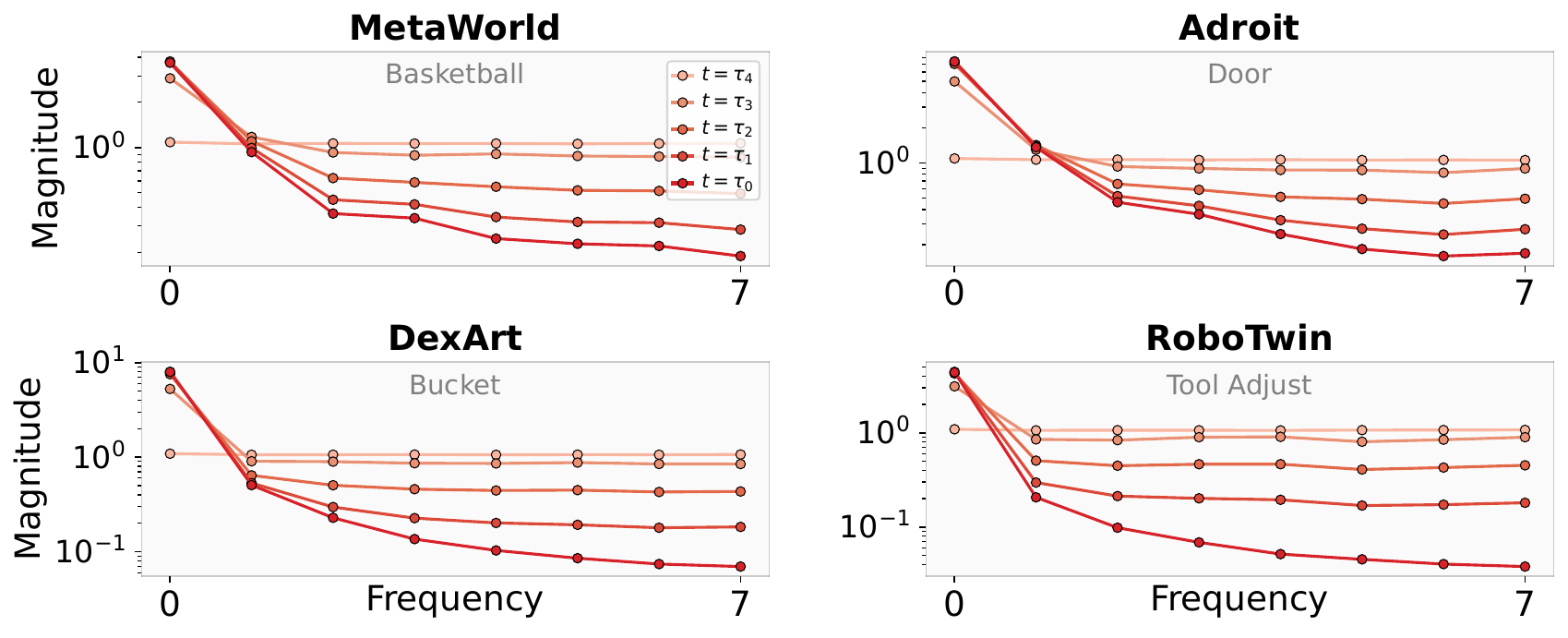}
    \caption{\textbf{Action DFT results.} As the denoising process progresses, the Gaussian noise ($t=\tau_{4}$) is gradually transformed into the predicted action ($t=\tau_{0}$). Timesteps $\tau_{i}$ is arranged in descending order of noise levels. The results reveal a consistent frequency evolution pattern: low-frequency components predominantly emerge during the early stages of denoising, whereas high-frequency features are progressively introduced in the latter phases of the process.}
    \label{fig:dft}
\end{wrapfigure}
To gain a more comprehensive understanding of the action generation, we apply Discrete Fourier Transform~(DFT) to examine how the frequency composition of actions evolves throughout the denoising process. Specifically, we conduct the analysis across $4$ benchmarks and visualize the spectral characteristics of action chunks during generation. As shown in Figure~\ref{fig:dft}, the results consistently indicate that the denoising process begins with the synthesis of low-frequency features, which are incrementally complemented by higher-frequency features in later stages. This observation not only shows that action, akin to image, exhibits an intrinsic inductive bias in the diffusion process, but also elucidates the action generation mechanism of \ourshort, wherein actions are hierarchically composed to captured features across varying levels of granularity.

\begin{figure}[t]
  \centering
  \includegraphics[width=1.0\linewidth]{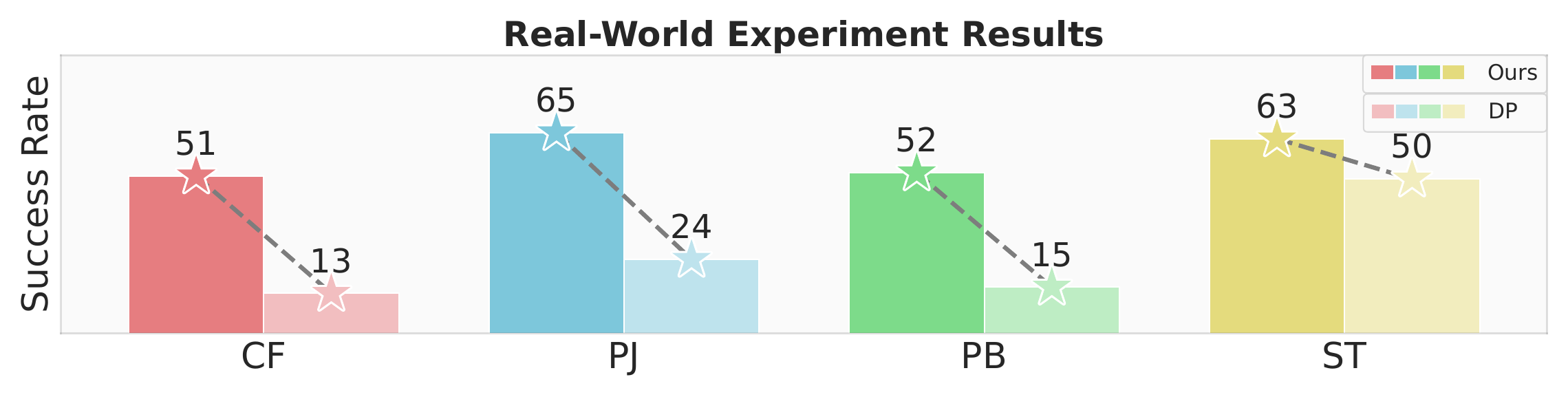}
  \caption{\textbf{\textbf{Success rate in real-world.}} Dark‑colored bars correspond to \ourshort, whereas the light‑colored bars correspond to DP. \ourshort outperforms DP in all 4 challenging real-world tasks.}
  \label{fig:realword-exp}
  \vspace{-10pt}
\end{figure}

\subsection{Real-world Experiments}

In terms of real-world experiments, we choose Galaxea R1 robot as our platform. We design four diverse challenging real-world tasks to evaluate the effectiveness of our method:\\
\textbf{Clean Fridge~(CF):} In a cluttered refrigerator environment, the robot is required to relocate a transparent bottle from the upper compartment to the lower one. The bottle is randomized within a 30 cm $\times$ 5 cm region on both the upper and lower shelves of the refrigerator. \\
\textbf{Pour Juice~(PJ):} This is a long-horizon task. The robot is required to place a cup in front of a water dispenser, scoop a spoonful of juice powder, then fill the cup with water, and finally put a straw in the cup. The cup is placed within a 7 cm $\times$ 7 cm area, and both the color of the juice powder and the position of the water dispenser are subject to variation across trials. \\
\textbf{Place Bottle~(PB):} The robot must place a bottle, initially located at a random position, onto a designated coaster. The bottle is placed within a 15 cm $\times$ 15 cm region, while the coaster is positioned within an around 25 cm $\times$ 25 cm area.\\
\textbf{Sweep Trash~(ST):} This long-horizon task entails picking up a broom, sweeping scattered debris on a table into a dustpan, and subsequently emptying the contents into a trash bin. The trash is randomly distributed across the entire table surface, approximately within a 40 cm $\times$ 40 cm area.

\begin{wrapfigure}{r}{0.45\textwidth}%
    \centering
    % \vspace{-23pt}
    \includegraphics[width=0.45\textwidth]{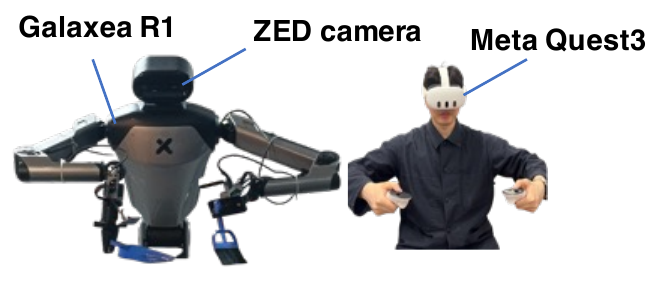}
    % \vspace{-15pt}
    \caption{\textbf{Experiment Setup. }}
    \label{fig:setup}
\end{wrapfigure}

\subsubsection{Experiment Setup}

We use the ZED camera to acquire the depth image with 60Hz running frequency. The demonstrations are collected by Meta Quest3. Regarding the two long-horizon tasks, both the baseline and our method incorporate the pre-trained ResNet18~\cite{he2016deep} encoders for RGB modality to enhance the policy’s perceptual capabilities in real-world environments. Each task is evaluated at 20 randomly sampled positions within the defined randomization range for each method. We record the success trials and calculate the corresponding success rate. In addition, during policy deployment, we adopt an \textit{asynchronous} design to obtain an approximately 15Hz inference speed. We also introduce \textit{temporal ensembling} and \textit{p-masking} to improve temporal consistency and alleviate overfitting to the proprioception state. More setup details can be found in Appendix~\ref{app:realworld}. 

\subsubsection{Experiment Results}

\textbf{Spatial generalization:} As shown in Figure~\ref{fig:realword-exp}, \ourshort significantly outperforms the baseline across all four real-world tasks, achieving an average improvement of $\mathbf{+32.3\%}$. It should be noted that in \textbf{CF} and \textbf{PJ} tasks, the policy is required to not only identify target objects in cluttered visual environments but also perform long-horizon reasoning to accomplish the tasks. While DP struggles to complete either task, \ourshort achieves substantial improvements, outperforming DP by $\mathbf{+38\%}$ and $\mathbf{+41\%}$ respectively. Therefore, \ourshort demonstrates superior perceptual and decision-making capabilities compared to alternative algorithms. Meanwhile, it should be noted that in terms of the point cloud based method DP3, it requires precise segmentation and high-fidelity depth sensing, resulting in it being less effective in handling our four cluttered real-world scenes that we designed. Comparative experimental results for DP3 are presented in Appendix~\ref{app:dp3realworld}.

\begin{table}[t]
\centering
\caption{\textbf{Instance generalization results.} \ourshort achieves $+15.4\%$ performance gain. }
\label{table: instance general}
\resizebox{0.9\textwidth}{!}{%
\begin{tabular}{l|ccc|ccc|cc}
\toprule

\begin{tabular}[c]{@{}c@{}} \multicolumn{1}{c}{\multirow{2}{*}{Method $\backslash$ Tasks}} \end{tabular}   & \multicolumn{3}{c|}{Place Bottle} & \multicolumn{3}{c|}{Sweep Trash} &  \multirow{2}{*}{\textbf{Average}} \\
 & coke bottle & sprite & can & $8\,\text{cm}^3$ & $64\,\text{cm}^3$ & $216\,\text{cm}^3$  & \\

\midrule
\bfourshort & \singleddbf{67} & \singleddbf{49} & \singleddbf{53} & \singleddbf{75} & \singleddbf{86} & \singleddbf{67} & \ccbf{66.2}\\
Diffusion Policy &	$45$ & $36$ & $40$ & $52$ & $72$ & $60$ & $50.8$ \\
\bottomrule
\end{tabular}}
\vspace{-1pt}
\end{table}

\textbf{Instance generalization:} 
Regarding instance generalization, we evaluate the model on two real-world tasks by varying the size and shape of bottles or trash items. As shown in Table~\ref{table: instance general}, after replacing the objects with variants of differing sizes and shapes, \ourshort maintains strong generalization capabilities attributable to its ability to hierarchically model features at multiple levels of granularity, and consistently outperforms baseline approaches across all settings.

\subsection{Ablation Study}
In this section, we ablate each key component of our framework and conduct experiments on three benchmarks to further exhibit the effectiveness of \ourshort.  The entire results in each benchmark can be found in Appendix \ref{app:ablation_whole}. 

\begin{wraptable}[8]{r}{0.6\textwidth}
\centering
\vspace{-0.18in}
\caption{\textbf{Ablation on hierarchical features.}}
\label{table:ab_h}
\vspace{-0.05in}
\resizebox{0.6\textwidth}{!}{%
\begin{tabular}{l|ccc|ccc}
\toprule
Methods $\backslash$ Benchmarks & MW & MS  & RT &  \textbf{Average} \\
\midrule
\bfourshort & \singleddbf{65.7} & \singleddbf{68.0} & \singleddbf{45.0}  & \ccbf{59.6} \\
w/o depth layering & $55.0$ & $52.5$  & $32.0$ & $46.5$ \\
w/o hierarchical action & $57.0$ & $50.0$  & $40.0$ & $49.0$  \\
w/o multi-scale representation & $53.7$ & $52.5$  & $40.0$ & $48.7$ \\
DP~(w/ depth) & $46.7$ & $47.5$  & $32.0$ & $42.1$  \\
\bottomrule
\end{tabular}}
\end{wraptable}

\textbf{Each hierarchical design.} We ablate the three hierarchical components introduced in our framework and compare them against the baseline Diffusion Policy with RGB-D input. As shown in Table~\ref{table:ab_h}, each hierarchical component independently contributes to performance improvement, consistently outperforming the DP~(w/ depth). Furthermore, Table~\ref{table:ab_h} also demonstrates that the integration of all three hierarchical designs leads to a substantial enhancement in overall performance.

\begin{wraptable}{r}{0.5\textwidth}
\centering
\caption{\textbf{Ablation on number of layers $\boldsymbol N$.}}
\label{table:ab_n}
\resizebox{0.5\textwidth}{!}{%
\begin{tabular}{c|ccc|ccc}
\toprule
Methods $\backslash$ Benchmarks & MW & MS  & RT &  \textbf{Average} \\
\midrule
\ourshort ($N=1$) & $55.0$ & $52.5$ & $32.0$  & $46.5$ \\
\ourshort ($N=2$) & $55.7$ & $60.0$  & $35.0$ & $50.2$ \\
\ourshort (\textbf{$N=3$}) & $65.7$ & \singleddbf{68.0} & $45.0$  & \ccbf{59.6} \\
\ourshort (\textbf{$N=4$})  & \singleddbf{67.0} & $61.5$ & \singleddbf{50.0}  & \ccbf{59.5} \\
\ourshort ($N=5$) & $58.7$ & $55.0$ & \singleddbf{50.0}  & $54.6$ \\
\ourshort ($N=6$) & $56.0$ & $51.0$ & $40.0$  & $49.0$ \\
\bottomrule
\end{tabular}}
\end{wraptable}

\textbf{The choice of $\boldsymbol N$ in depth-aware layering.} For the depth-aware layering component, we investigate whether the policy’s performance is sensitive to the choice of the number of layers $N$. As presented in Table ~\ref{table:ab_n}, the trained policy achieves optimal and comparable performance when $N$ is set to 3 or 4, a trend consistently observed across all evaluated benchmarks. When $N$ becomes excessively large, the image is over-partitioned, thus reducing the representation capacity of the policy. Nevertheless, even in such cases, the performance remains slightly better than the non-layered baseline. These findings highlight the critical role of depth-aware layering in enhancing policy effectiveness.

\section{Conclusion}
In this paper, we introduce \ourshort, an efficient generalizable visuomotor policy learning framework that can obtain superior performance in a wide range of simulations and challenging real-world tasks. Extensive empirical evidence suggests that establishing a more cohesive integration between visual feature representations and the action generation process can enhance the generalization capacity and learning efficiency of our learned policies. The proposed three hierarchical designs not only facilitate the effective fusion of RGB and depth modalities, but also strengthen the correspondence between visual features and the generated actions at different granularity levels. In the future, we expect to extend the applicability of \ourshort to more intricate and fine-grained dexterous real-world tasks.
\section{Limitations}
Although \ourshort has demonstrated effectiveness in a variety of tasks, there exist several limitations. First, despite our use of asynchronous execution to improve inference speed in real-world settings, the overall inference time of diffusion-based models remains relatively slow. We could explore distilling the policy into a consistency model, to enhance real-time performance. Second, the limited depth quality of the ZED camera may hinder the policy’s full potential in real-world deployment; employing higher-fidelity depth sensors could further boost the effectiveness of \ourshort in practical scenarios.

%===============================================================================
\acknowledgments{We are thankful to all members of Galaxea for helping with hardware infrastructure and real-world experiments. We also thank the members of TEA Lab for their helpful discussions.}
	
\clearpage
% The acknowledgments are automatically included only in the final and preprint versions of the paper.
%===============================================================================

\bibliography{reference}  % .bib

\newpage
\appendix
\section*{Appendix}
\section{Hyperparameters}
\label{app:hyper}

To effectively address the varying levels of difficulty and distinct properties inherent to different benchmarks, we adapt our hyperparameter settings for each specific dataset. The chosen configurations, detailed in Table~\ref{table:hyperparams0}, \ref{table:hyperparams1}, \ref{table:hyperparams2}, \ref{table:hyperparams3}, are selected based on previous works~\cite{chi2023diffusion, Ze2024DP3, zhu2024point, mu2024robotwin}.

\begin{table}[ht]
\centering
\caption{\textbf{Hyperparameters used for MetaWorld, DexArt.}}
\begin{tabular}{l|c}
\toprule
\textbf{Hyperparameter} & \textbf{Value} \\
\midrule
Observation Horizon ($T_o$) & 2 \\
Action Horizon ($T_a$) & 2 \\
Prediction Action Horizon ($T_p$) & 4 \\
Optimizer & AdamW~\cite{kingma2014adam} \\
Betas ($\beta_1,\beta_2$) & [0.95, 0.999] \\
Learning Rate & 1.0e-4 \\
Weight Decay & 1.0e-6 \\
Learning Rate Scheduler & Cosine\\
Training Timesteps ($T$) & 50 \\
Inference Timesteps & 20 \\
Prediction Type & $\epsilon$-prediction\\
Image Resolution & 128 $\times$ 128 \\
Scale Number ($K$) & 4\\
Multi-Scale Representation Resolutions ($\{(h_k,w_k)\}_{k=1}^{K}$) & \{(1,1),(3,3),(5,5),(7,7)\} \\
Stage Boundiaries ($\{\tau_k / T\}_{k=0}^K$) & \{0,0.4,0.6,0.8,1.0\}\\
% Noise Scheduler & DDIM
\bottomrule
\end{tabular}
\label{table:hyperparams0}
\end{table}

\begin{table}[ht]
\centering
\caption{\textbf{Hyperparameters used for Adroit.}}
\begin{tabular}{l|c}
\toprule
\textbf{Hyperparameter} & \textbf{Value} \\
\midrule
Observation Horizon ($T_o$) & 2 \\
Action Horizon ($T_a$) & 2 \\
Prediction Action Horizon ($T_p$) & 4 \\
Optimizer & AdamW\\
Betas ($\beta_1,\beta_2$) & [0.95, 0.999] \\
Learning Rate & 1.0e-4 \\
Weight Decay & 1.0e-6 \\
Learning Rate Scheduler & Cosine\\
Training Timesteps ($T$) & 50 \\
Inference Timesteps & 20 \\
Prediction Type & $\epsilon$-prediction\\
Image Resolution & 84 $\times$ 84 \\
Scale Number ($K$) & 4\\
Multi-Scale Representation Resolutions ($\{(h_k,w_k)\}_{k=1}^{K}$) & \{(1,1),(3,3),(5,5),(6,6)\}\\
Stage Boundiaries ($\{\tau_k / T\}_{k=0}^K$) & \{0,0.4,0.6,0.8,1.0\}\\
% Noise Scheduler & DDIM
\bottomrule
\end{tabular}
\label{table:hyperparams1}
\end{table}

\begin{table}[ht]
\centering
\caption{\textbf{Hyperparameters used for ManiSkill.}}
\begin{tabular}{l|c}
\toprule
\textbf{Hyperparameter} & \textbf{Value} \\
\midrule
Observation Horizon ($T_o$) & 2 \\
Action Horizon ($T_a$) & 8 \\
Prediction Action Horizon ($T_p$) & 16 \\
Optimizer & AdamW\\
Betas ($\beta_1,\beta_2$) & [0.9, 0.95] \\
Learning Rate & 1.0e-4 \\
Weight Decay & 1.0e-4 \\
Learning Rate Scheduler & One Cycle LR~\cite{smith2019super} \\
Training Timesteps ($T$) & 100 \\
Inference Timesteps & 100 \\
Prediction Type & $\epsilon$-prediction \\
Image Resolution & 128 $\times$ 128 \\
Scale Number ($K$) & 4\\
Multi-Scale Representation Resolutions ($\{(h_k,w_k)\}_{k=1}^{K}$) & \{(1,1),(3,3),(5,5),(7,7)\}\\
Stage Boundaries ($\{\tau_k\}_{k=0}^K / T$) & \{0,0.4,0.6,0.8,1.0\}\\
% Noise Scheduler & DDPM
\bottomrule
\end{tabular}
\label{table:hyperparams2}
\end{table}

\begin{table}[ht]
\centering
\caption{\textbf{Hyperparameters used for RoboTwin.}}
\begin{tabular}{l|c}
\toprule
\textbf{Hyperparameter} & \textbf{Value} \\
\midrule
Observation Horizon ($T_o$) & 3 \\
Action Horizon ($T_a$) & 2 \\
Prediction Action Horizon ($T_p$) & 8 \\
Optimizer & AdamW\\
Betas ($\beta_1,\beta_2$) & [0.95, 0.999] \\
Learning Rate & 1.0e-4 \\
Weight Decay & 1.0e-6 \\
Learning Rate Scheduler & Cosine \\
Training Timesteps ($T$) & 100 \\
Inference Timesteps & 100 \\
Prediction Type & $\epsilon$-prediction \\
Image Resolution & 180 $\times$ 320 \\
Scale Number ($K$) & 4\\
Multi-Scale Representation Resolutions ($\{(h_k,w_k)\}_{k=1}^{K}$) & \{(1,3),(3,5),(5,7),(5,9)\}\\
Stage Boundaries ($\{\tau_k\}_{k=0}^K / T$) & \{0,0.4,0.6,0.8,1.0\}\\
% Noise Scheduler & DDPM
\bottomrule
\end{tabular}
\label{table:hyperparams3}
\end{table}

In addition to the hyperparameters reported in the table, the choice of the number of layers $N$ demonstrates great importance, as shown in Table~\ref{table:ab_n}. Empirically, we choose $N=4$ in Adroit, MetaWorld Hard and Hard++, and $N=3$ in other benchmarks. 

The noise scheduler for diffusion process is determined by $\alpha_t$, defined using function $f(t)$
\begin{equation}
    \alpha_t = \frac{f(t)}{f(0)},\quad\text{where}\quad f(t) = \cos^2\left(\frac{\pi}{2}\frac{t/T+s}{1+s}\right).
\end{equation}
Here, $T$ is the total number of diffusion timesteps and $s$ is an offset parameter.

For the reverse process, we employ different formulations depending on the environment. In MetaWorld, Adroit and DexArt, we follow the DDIM~\cite{song2020denoising} approach, formulating the reverse process as an ODE, which corresponds to setting
\begin{equation}
    \sigma_t = 0
\end{equation}
for all $t$. In ManiSkill and RoboTwin, we follow the design of DDPM~\cite{ho2020denoising} and formulate the reverse process as a Variance Preserving (VP) SDE~\cite{song2020score}. In this case, for all $t$,
\begin{equation}
    \sigma_t = \sqrt{\frac{1-\alpha_{t-1}}{1-\alpha_t}}\sqrt{1-\frac{\alpha_t}{\alpha_{t-1}}}.
\end{equation}

Furthermore, consider the weighting term $\gamma_t$ in Equation~\ref{eq:diffloss}. Since the choice of $\gamma_t$ does not affect the optimal denoising network $\epsilon_{\theta^*}$, we set
\begin{equation}
    \gamma_t = 1
\end{equation}
for all $t$.

\section{Method Details}

This section outlines the implementation details of our multi-scale encoding. The encoder $\mathcal{E}_m$ for each depth layer $m$ adopts the architecture from VQGAN~\cite{esser2021taming}, ensuring strong representational capacity while preserving spatial information. We use $\operatorname{interpolate}$ to denote a differentiable resizing operation (e.g. bilinear or nearest-neighbor interpolation), which is crucial for enabling gradient flow during training. The function $\mathcal{Q}$ represents the quantization process detailed in Equation~\ref{eq:quant}. Finally, after interpolating a feature map $f_{m,k}$ to the highest resolution, we apply a lightweight convolutional network $\phi_{m,k}$ designed to help restore fine details from the potentially lower-resolution source features.

The pseudocode for this process is outlined in Algorithm~\ref{alg:enc}.
\label{app:method_details}
\begin{center}
  \centering
  \begin{algorithm}[H]
    \caption{Multi-scale Encoding} \label{alg:enc}
    \textbf{Inputs: } raw image $I$\\
    \textbf{Hyperparameters: } depth layer number $N$, scale number $K$, resolutions $\{(h_k,w_k)\}_{k=1}^{K}$\\
    Partition image $I$ into $N+1$ images $\{I_m\}_{m=0}^N$ according to Equation~\ref{eq:layer}\\
    \For {$m = 0, \dots, N-1 $}
    {
    $f_m\leftarrow\mathcal E_m(I_m)\in \mathbb R^{h_K\times w_K\times C}$\\
    \For {$k=1,\dots,K$}
    {
    $f_{m,k} \leftarrow \operatorname{interpolate}(f_m, h_k, w_k)\in \mathbb R^{h_k\times w_k\times C}$\\
    $f_{m,k} \leftarrow \mathcal{Q}(f_{m,k})$\\
    $f_{m,k} \leftarrow \phi_{m,k}({\operatorname{interpolate}(f_{m,k}, h_K, w_K)})\in \mathbb R^{h_K\times w_K\times C}$\\
    $\hat{f}_{m,k}\leftarrow\sum_{k'\leq k} f_{m,k'}$\\
    $f_m \leftarrow f_m - f_{m,k}$
    }
    }
    \textbf{Return: } multi-scale features $F=\{\hat{f_k}=\{\hat{f}_{m,k}\}_{m=0}^{N-1}\}_{k=1}^{K}$
  \end{algorithm}
\end{center}

All trainable parameters, including the visual encoders $\{\mathcal{E}_m\}_{m=0}^{N-1}$, the codebooks $\{\mathcal{Z}_m\}_{m=0}^{N-1}$, the CNN parameters $\{\{\phi_{m,k}\}_{m=0}^{N-1}\}_{k=1}^{K}$, and the denoising network $\epsilon_\theta$, are trained jointly in an end-to-end manner. The optimization minimizes the combined objective function $\mathcal{L}$, defined as a weighted sum of consistency loss (Equation~\ref{eq:consistency}) and the diffusion loss (Equation~\ref{eq:diffloss}):
\begin{equation}
    \mathcal{L}= \mathcal{L}_{\text{diffusion}}+\alpha\mathcal{L}_{\text{consistency}},
\end{equation}

where $\alpha$ is a hyperparameter balancing the two loss terms.

\section{Expert Demonstrations}

\label{app:expert_demo}
Regarding the MetaWorld~\cite{yu2020meta} and the RoboTwin~\cite{mu2024robotwin} benchmarks, we utilize scripted policies to generate expert demonstrations. In the case of ManiSkill~\cite{gu2023maniskill2} tasks, we employ the officially provided demonstrations. Trajectories for other simulation benchmarks are collected with agents trained by RL algorithms~\cite{Ze2024DP3, schulman2017proximal, wang2022vrl3}. The expert policies are evaluated over 200 episodes, and their success rates are detailed in Table~\ref{table:expert}.

Given the varying difficulty levels across benchmarks, we provide a different number of demonstrations for each. Specifically, we provide 50 trajectories per task for MetaWorld, Adroit, and RoboTwin. For DexArt, we follow the setup in~\cite{Ze2024DP3} and provide 100 trajectories per task. For ManiSkill, we use all official demonstrations: 1000 for rigid tasks and 200 for deformable tasks.

In real-world experiments, we collect demonstrations of varying quantity, depending on the complexity and horizon length of the tasks. For short-horizon tasks, the number of collected trajectories is relatively limited --- 100 for Clean Fridge and 200 for Place Bottle. In contrast, long-horizon tasks demand more comprehensive data coverage. We collect more demonstrations: 270 for Pour Juice and 500 for Sweep Trash. These demonstrations play a crucial role in guiding the training process, especially in scenarios where exploration is challenging or unsafe.

\section{Real-world Training Details}

\label{app:realworld}

As mentioned in \cite{gong2024carp}, DP-based methods often suffer from low inference speed, which can cause the inference process to stall. Prior approaches, including DP3~\cite{Ze2024DP3}, attempt to address this by increasing action horizon (e.g. $T_a=4$ or $T_a=8$) or reducing the number of model parameters (e.g. Simple DP3). However, these strategies often compromise manipulation accuracy and dexterity. A further complication is that increasing $T_a$ widens the temporal gap between consecutive inference steps, leading to greater discrepancies in observed information, and consequently, divergence in predicted actions. This often results in noticeable jitter and discontinuities in manipulation.

In general, DP-based methods are hindered by low inference speed, temporal inconsistency and overfitting to proprioceptive information. To address these challenges and improve real-world performance, we employ several empirical techniques.

\subsection{Higher Inference Speed}

\label{app:higherinference}

To mitigate slow inference rooted in DP, we adopt an \textit{asynchronous} design, achieving a final inference frequency of $10$-$15$ Hz. Instead of waiting for the execution of all predicted actions before initiating the next inference cycle, our method performs inference concurrently with action execution. The predicted action is stored in a queue to be executed at a fixed inference speed ($10$-$15$ Hz in practice, $12$ Hz as average).

The inference speeds achieved in real-world scenarios are presented in Table~\ref{table:real_infer}. \ourshort (asynchronous) demonstrates a superior inference speed compared to standard DP~\cite{chi2023diffusion} and DP3~\cite{Ze2024DP3}, as well as our synchronous \ourshort implementation. In addition to this speed advantage, \ourshort features a shorter action sequence length ($T_a=2$), which contributes to more dexterous manipulation capabilities.

\begin{table}[ht]
\centering
\renewcommand\tabcolsep{8.0pt}
    \caption{\textbf{Comparison of real-world inference speeds for different methods.} The asynchronous version of our method demonstrates a significant speed-up by decoupling inference from action execution.}
    \begin{tabular}{lccccc}
        \toprule
        {Method} & DP & DP3 & \ourshort & \ourshort (asynchronous)\\
        \midrule
        Inference Speed (FPS) & 12.4 & 12.7 & 12.1 & \best{24.2}\\
        \bottomrule
    \end{tabular}
    \label{table:real_infer}
\end{table}

\subsection{Temporal Consistency}

Having adopted the \textit{asynchronous} design, we have obtained action sequences with overlapping time intervals. To ensure temporal smoothness and reduce discontinuities, we incorporate \textit{temporal ensembling mechanism} from ACT~\cite{zhao2023learning}. As in ACT, \ourshort performs a weighted average of actions with the same timestep across multiple overlapping sequences. This ensembling mitigates the gap between actions inferred from slightly different observations and effectively reduces jitter.

\subsection{Alleviate Overfitting}

Similar to other real-world robotic systems, \ourshort is susceptible to overfitting on proprioceptive inputs, often neglecting the RGB-D information. This is evidenced by that the model generates similar actions regardless of variations in object positions. We hypothesize that this occurs because the simple, low-parameter MLP used to encode proprioception is easier to optimize than the more complex CNN used for RGB-D input, leading to reliance on the former.

To mitigate this, we introduce a \textit{p-masking} strategy during training. This mechanism stochastically masks all proprioceptive inputs with probability $p$, which decays linearly over the training process. Specifically, for training timestep $t$ in a total horizon $T$, $p(t)=1-t/T$. This schedule encourages the model to rely more on RGB-D features early in training, helping it avoid early-stage overfitting and develop stronger visual grounding. 

\section{Additional Experiment Results}

\subsection{Simulation Results for Each Task}
\label{app:sim_total}

We present the simulation results for each task in Table~\ref{table:sim_results_total}, which serves as a supplement to Table~\ref{table:sim-results}. For each experiment, we report the average success rate over three different random seeds. The final average result is obtained by averaging across all benchmarks.

\begin{figure}[ht]
  \centering
  \includegraphics[width=1.0\linewidth]{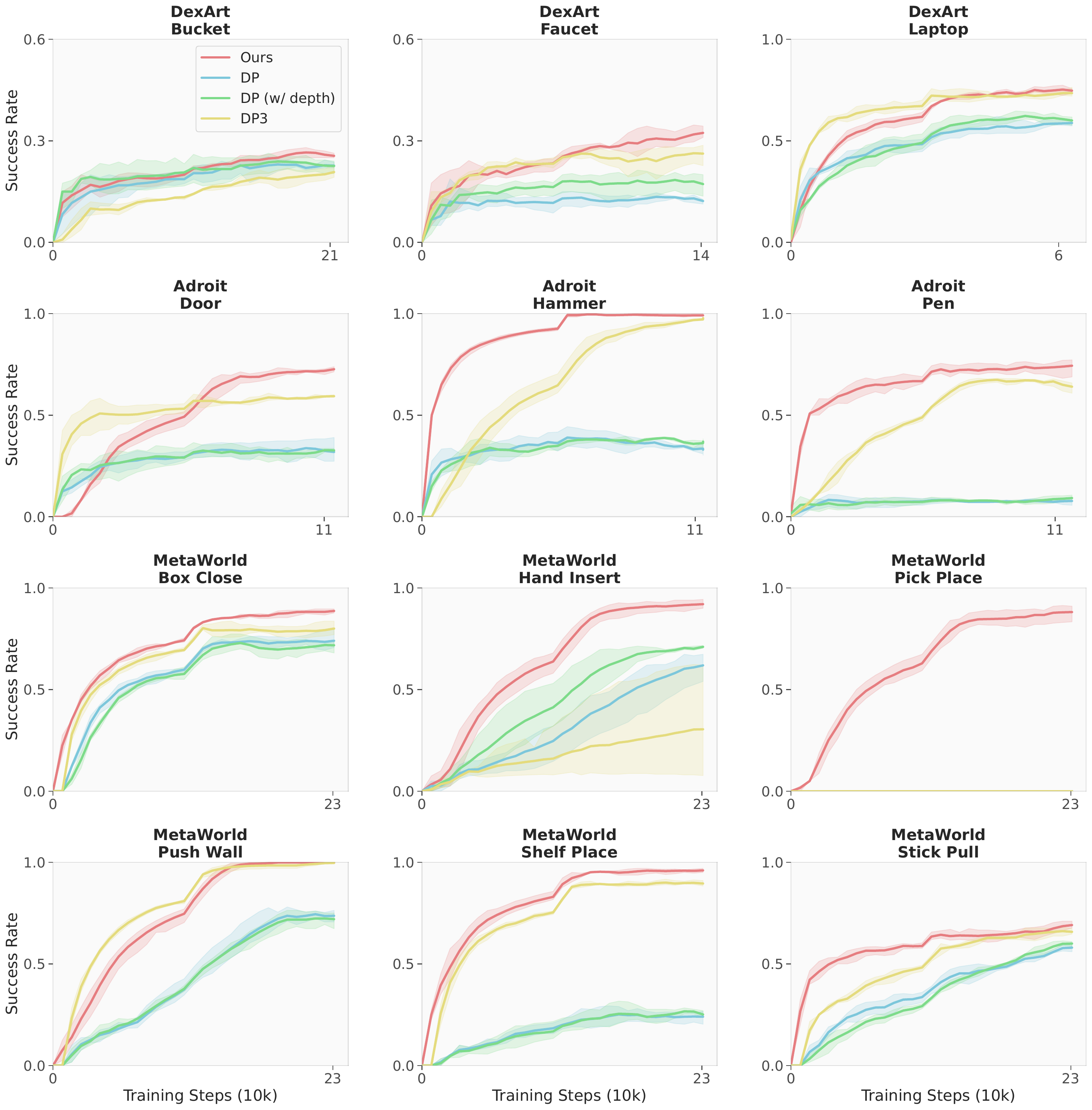}
  \caption{\textbf{Learning curves of the four methods on 12 randomly sampled diverse simulation tasks.} In most tasks, \ourshort demonstrates faster convergence, higher final success rates, and lower variance compared to other three methods.}
  \label{fig:train_curve}
\end{figure}

We also provide the training progress of 4 algorithms on 12 various tasks across 3 different benchmarks in Figure~\ref{fig:train_curve}. The selected tasks span a range of difficulties and are included without cherry picking to provide an unbiased view of each algorithm. 

\subsection{The Whole Results of Ablation Study}
\label{app:ablation_whole}

We present the entire results of our ablation study on each hierarchical design and number of layers $N$ in Table~\ref{table:ablation_h_total} and Table~\ref{table:ablation_N_total}, as a supplement to Table~\ref{table:ab_h} and Table~\ref{table:ab_n}. For each experiment, the success rate is reported by averaging over 3 different random seeds. The final average result is obtained by averaging across benchmarks.

\begin{table}[ht]
\centering
\caption{\textbf{Whole results of ablation study on hierarchical features.}}
\label{table:ablation_h_total}
\resizebox{\textwidth}{!}{
\begin{tabular}{l|ccc|cc|c|c}
\toprule
\begin{tabular}[c]{@{}c@{}} \multicolumn{1}{c}{\multirow{2}{*}{Method  $\backslash$ Tasks}} \end{tabular} & \multicolumn{3}{c|}{\textbf{MetaWorld}} & \multicolumn{2}{c|}{\textbf{ManiSkill}} & \multicolumn{1}{c|}{\textbf{RoboTwin}} & \multicolumn{1}{c}{\multirow{2}{*}{\textbf{Average}}} \\
 & \small Soccer & \small Stick Pull &\small Pick Out of Hole &\small Fill &\small Excavate &\small Tool Adjust & \\
\midrule
\bfourshort & \best{85} & \best{75} & 37 & \best{98} & \best{38} & \best{45} & \best{59.6}\\
w/o depth layering & 59 & 72 & 34 & 78 & 27 & 32 & {46.5}\\
w/o hierarchical action & 64 & 67 & \best{40} & 82 & 18 & 40 & {49.0}\\
w/o multi-scale representation & 55 & 72 & 34 & 73 & 32 & 40 & 48.7\\
DP~(w/ depth) & 37 & 71 & 32 & 72 & 23 & 32 & {42.1}\\
\bottomrule
\end{tabular}}

\end{table}

\begin{table}[ht]
\centering
\caption{\textbf{Whole results of ablation study on number of layers $\boldsymbol N$.}}
\label{table:ablation_N_total}
\resizebox{\textwidth}{!}{
\begin{tabular}{c|ccc|cc|c|c}
\toprule
\begin{tabular}[c]{@{}c@{}} \multicolumn{1}{c}{\multirow{2}{*}{Method  $\backslash$ Tasks}} \end{tabular} & \multicolumn{3}{c|}{\textbf{MetaWorld}} & \multicolumn{2}{c|}{\textbf{ManiSkill}} & \multicolumn{1}{c|}{\textbf{RoboTwin}}  & \multicolumn{1}{c}{\multirow{2}{*}{\textbf{Average}}}\\
 & \small Soccer & \small Stick Pull &\small Pick Out of Hole &\small Fill &\small Excavate &\small Tool Adjust & \\
\midrule
\ourshort ($N=1$) & 59 & 72 & 34 & 78 & 27 & 32 & 46.5\\
\ourshort ($N=2$) & 64 & 70 & 33 & 85 & 35 & 35 & 50.2\\
\ourshort ($N=3$) & \best{85} & 75 & 37 & \best{98} & \best{38} & 45 & \best{59.6}\\
\ourshort ($N=4$) & 78 & \best{83} & \best{40} & 90 & 33 & \best{50} & \best{59.5}\\
\ourshort ($N=5$) & 62 & 75 & 39 & 87 & 23 & \best{50} & 54.6\\
\ourshort ($N=6$) & 61 & 73 & 34 & 77 & 25 & 40 & 49.0\\
\bottomrule
\end{tabular}}

\end{table}

\subsection{Comparison with a GMM-based Layering Variant}
\label{app:gmm}

To highlight the advantages of depth-aware layering, we conduct a comparison against a variant where this module is substituted with a classical foreground-background segmentation method, Gaussian Mixture Models~(GMM)~\cite{reynolds2009gaussian}, named \ourshort-GMM. As shown in Table~\ref{table:gmm}, \ourshort outperforms \ourshort-GMM in all benchmarks. Notably, \ourshort-GMM yields results comparable to a simple single-layer ($N=1$) approach, further emphasizing the rationality and effectiveness of our proposed depth-aware layering strategy.
\begin{table}[ht]
\centering
\caption{\textbf{Comparison with GMM-based layering variant.} \ourshort with depth-aware layering achieves superior performance compared to using GMM for layering.}
\label{table:gmm}
\resizebox{\textwidth}{!}{
\begin{tabular}{c|ccc|cc|c|c}
\toprule
\begin{tabular}[c]{@{}c@{}} \multicolumn{1}{c}{\multirow{2}{*}{Method  $\backslash$ Tasks}} \end{tabular} & \multicolumn{3}{c|}{\textbf{MetaWorld}} & \multicolumn{2}{c|}{\textbf{ManiSkill}} & \multicolumn{1}{c|}{\textbf{RoboTwin}} & \multicolumn{1}{c}{\multirow{2}{*}{\textbf{Average}}} \\
 & \small Soccer & \small Stick Pull &\small Pick Out of Hole &\small Fill &\small Excavate &\small Tool Adjust\\
\midrule
\bfourshort & \best{85} & \best{83} & \best{40} & \best{98} & \best{38} & \best{45} & \best{64.8} \\
\ourshort-GMM & 45 & 67 & 32 & 75 & 27 & 37 & 47.2\\
\ourshort ($N = 1$) & 59 & 72 & 34 & 78 & 27 & 32 & 50.3\\
\bottomrule
\end{tabular}}

\end{table}

\subsection{Comparison with More Baselines}

Except diffusion-based algorithms, we also compare \ourshort with the recent state-of-the-art method CARP~\cite{gong2024carp}, which uses multi-scale action VQ-VAE to build hierarchical action structures. Table~\ref{table:h3dp_carp} shows that \ourshort outperforms CARP with an average improvement of 18.9\%, indicating the importance of adopting hierarchical designs throughout visual features and action generation. 

\begin{table}[ht]
\centering
\caption{\textbf{Comparison with CARP.} \ourshort outperforms CARP with an average improvement of 18.9\%.}
\label{table:h3dp_carp}
\resizebox{\textwidth}{!}{
\begin{tabular}{c|ccccccc|c}
\toprule
\begin{tabular}[c]{@{}c@{}} \multicolumn{1}{c}{\multirow{2}{*}{Method  $\backslash$ Tasks}} \end{tabular} & \multicolumn{7}{c|}{\textbf{MetaWorld}} & \multicolumn{1}{c}{\multirow{2}{*}{\textbf{Average}}} \\
 & \small Box Close & \small Soccer &\small Stick Pull &\small Pick Out of Hole &\small Peg Insert Side &\small Hammer &\small Sweep\\
\midrule
\ourshort & \best{98} & \best{85} & \best{83} & \best{40} & \best{98} & \best{100} & \best{100} & \best{86.3}\\
CARP & 82 & 53 & 71 & 15 & 69 & 82 & \best{100} & 67.4\\
DP & 83 & 43 & 64 & 13 & 62 & 64 & 96 & 60.7\\
\bottomrule
\end{tabular}}

\end{table}

\subsection{Comparison with DP with Pre-trained Visual Encoder}

Prior work suggests that pre-trained visual representation may enhance spatial generalization of policy~\cite{xue2025demogen}. Hence, we investigate the impact of integrating a pre-trained visual encoder with the original DP. We specifically replace the standard ResNet encoder in DP with DINOv2~\cite{oquab2023dinov2} model. 

This variant, named DP-DINOv2, is evaluated on randomly selected tasks from the MetaWorld benchmark. The comparative results are presented in Table~\ref{table:dino}. Although DP-DINOv2 shows a marginal improvement on some tasks compared to the original DP baseline, this comes with drawbacks, including a longer training time, inference latency and larger number of parameters($\sim$21M for DINOv2 with ViT-S) due to the DINOv2 architecture. 

In contrast, \ourshort utilizes an efficient visual encoder with less than 0.7M parameters, which achieves strong performance improvements over the original DP without incurring the aforementioned overheads.

\begin{table}[ht]
\centering
\caption{\textbf{Comparison with DP with pre-trained visual encoder.} While DP-DINOv2 yields small improvement after paying additional cost, \ourshort demonstrates superior performance.}
\label{table:dino}
\resizebox{\textwidth}{!}{
\begin{tabular}{c|cccccc|c}
\toprule
\begin{tabular}[c]{@{}c@{}} \multicolumn{1}{c}{\multirow{2}{*}{Method  $\backslash$ Tasks}} \end{tabular} & \multicolumn{6}{c|}{\textbf{MetaWorld}} & \multicolumn{1}{c}{\multirow{2}{*}{\textbf{Average}}} \\
 & \small Hand Insert & \small Pick Out of Hole &\small Disassemble &\small Stick Pull &\small Soccer &\small Sweep Into\\
\midrule
\bfourshort & \best{100} & \best{40} & \best{96} & \best{83} & \best{85} & \best{100} & \best{84.0}\\
DP & 73 & 13 & 81 & 64 & 43 & 74 & 58.0\\
DP-DINOv2 & 91 & 24 & 77 & 72 & 41 & 78 & 63.8\\
\bottomrule
\end{tabular}}

\end{table}

\subsection{Importance of Segmentation in DP3}
\label{app:segdp3}

As highlighted in Section~\ref{sec:sim_per}, DP3 relies on manual segmentation of point cloud for optimal performance. To demonstrate this dependency, we evaluate DP3's performance under two distinct segmentation conditions using randomly selected tasks from the MetaWorld benchmark.

We compare the following two scenarios: \textit{DP3 with ideal segmentation}, which utilizes clean segmented point clouds containing only the robot and task-relevant objects, as implemented in the original DP3 algorithm; \textit{DP3 without ideal segmentation}, which utilizes point clouds that are intentionally processed to include desk surface upon which objects rest, while other background elements are still removed. This configuration simulates common real-world scenarios where simple or automated segmentation rules might fail to perfectly isolate the task-relevant objects.

As shown in Table~\ref{table:dp3_seg}, DP3's performance degrades substantially when operating on point clouds without ideal segmentation. This result confirms that DP3 is highly sensitive to the quality of the input point cloud segmentation.

\begin{table}[t]
\centering
\caption{\textbf{Comparision of DP3 under different segmentation qualities.} We compare DP3 success rates on selected tasks when provided with different segmentation qualities, highlighting significant performance degradation.}
\label{table:dp3_seg}
\resizebox{\textwidth}{!}{
\begin{tabular}{c|cccccc|c}
\toprule
\begin{tabular}[c]{@{}c@{}} \multicolumn{1}{c}{\multirow{2}{*}{Method  $\backslash$ Tasks}} \end{tabular} & \multicolumn{6}{c|}{\textbf{MetaWorld}} & \multicolumn{1}{c}{\multirow{2}{*}{\textbf{Average}}} \\
 &\small Push & \small Shelf Place &\small Stick Pull &\small Soccer &\small Bin Picking &\small Pick Place Wall \\
\midrule
DP3 & \best{96} & \best{86} & \best{61} & \best{57} & \best{100} & \best{97} & \best{82.8}\\
DP3 (w/o ideal segmentation) & 89 & 26 & 48 & 29 & 50 & 84 & 54.3\\
\bottomrule
\end{tabular}}

\end{table}

In contrast, \ourshort operates directly on raw image without requiring such pre-processing, thereby avoiding such failure mode and the associated need for careful, potentially manual, segmentation tuning, especially common in real-world scenarios.

\subsection{\texorpdfstring{\bfourshort}{} in Tasks with Significant Depth Variations}
\label{app:sig_depth}

As introduced in Section \ref{sec:depthlayer}, our depth-aware layering mechanism discretizes the depth map into distinct layers. This layering offers a crucial advantage in scenarios with significant depth variations by providing a structured representation that preserves visual detail while emphasizing foreground-background separation. We will elaborate on this benefit and provide supporting comparative analysis here.

We conduct further experiments on tasks involving complex spatial arrangements, such as reaching for an object from closer to further or manipulating items in a cluttered scene, demanding a fine-grained understanding of relative object depth. Although raw RGB-D data contains this information implicitly, models may struggle to effectively utilize it, potentially treating the depth channel similarly to color channels or failing to prioritize significant depth discontinuities. Point cloud representations, inherently capturing 3D structures, often perform well in such scenarios as they directly encode geometric relationships.

Our depth-aware layering mechanism explicitly addresses this challenge for RGB-D inputs. By assigning pixels to discrete layers based on their depth values, we impose a structure that forces the model to differentiate between elements located at varying distances to camera. This discretization acts as an inductive bias, guiding the model to attend more strongly to the geometric layout and relative positioning of objects along the depth axis.

To empirically support our hypothesis, we conduct an ablation study focusing on tasks exhibiting significant depth variations. We compare the performance of three distinct approaches: DP3, DP~(w/ depth) and \ourshort (only with depth-aware layering, i.e., without hierarchical action and multi-scale representation).

As seen in Table \ref{table:h3dp_depth}, our observations reveal a consistent pattern: in tasks involving significant depth variations, point cloud--based policy initially demonstrated superior performance compared to standard RGB-D processing, represented by DP~(w/ depth). However, upon integrating the depth-aware layering mechanism, \ourshort consistently outperforms the baseline on these tasks, which strongly supports our claim.

\begin{table}[ht]
\centering
\caption{\textbf{Performance comparison demonstrating the effectiveness of depth-aware layering.} Tasks with significant depth variations show great improvement only with depth layering compared to DP (w/ depth), surpassing the point cloud baseline (DP3).}
\label{table:h3dp_depth}
\resizebox{\textwidth}{!}{
\begin{tabular}{c|cccccc|c}
\toprule
\begin{tabular}[c]{@{}c@{}} \multicolumn{1}{c}{\multirow{2}{*}{Method  $\backslash$ Tasks}} \end{tabular} & \multicolumn{6}{c|}{\textbf{MetaWorld}} & \multicolumn{1}{c}{\multirow{2}{*}{\textbf{Average}}} \\
 & \small Push & \small Shelf Place &\small Disassemble &\small Soccer &\small Pick Place Wall &\small Peg Insert Side\\
\midrule
\bfourshort~\textbf{(only w/ depth layering)} & \best{100} & \best{95} & \best{98} & 55 & \best{100} &  86 & \best{89.0} \\
% (only w/ depth layering)\\
DP (w/ depth) & 79 & 29 & 76 & 37 & 80 & 53 & 59.0\\
DP3 & 96 & 86 & \best{98} & \best{57} & 97 & \best{92} & 87.7\\
\bottomrule
\end{tabular}}

\end{table}

\subsection{Comparison with DP3 in Real-world Experiments}
\label{app:dp3realworld}

DP3 \cite{Ze2024DP3} is a renowned baseline succeeding DP \cite{chi2023diffusion} in imitation learning and robotic manipulation, achieving state-of-the-art results in multiple simulation environments. However, DP3 has notable limitations. In particular, it relies heavily on high-quality point clouds, typically requiring precision sensors such as the RealSense L515 to function effectively.

\begin{wraptable}{r}{0.5\textwidth}
\centering
\caption{\textbf{Comparison of \bfourshort and DP3 in real-world experiments.} We make comparison in $2$ short-horizon real-world tasks and both use LFS encoders. \ourshort achieves $+29.0\%$ performance gain. }
\label{table: h3dp_dp3}
% \resizebox{0.4\textwidth}{!}{%
\begin{tabular}{c|c|c|c}
\toprule

\begin{tabular}[c]{@{}c@{}} \multicolumn{1}{c}{\multirow{1}{*}{Method $\backslash$ Tasks}} \end{tabular}   & \multicolumn{1}{c|}{CF} & \multicolumn{1}{c|}{PB} &  \multirow{1}{*}{\textbf{Average}} \\

\midrule

\bfourshort & \best{51} & \best{52} & \best{51.5} \\

DP3 &	$12$ & $33$ & $22.5$ \\

\bottomrule
\end{tabular}
% }
\end{wraptable}

In our setup, the head-mounted camera is a ZED, which produces relatively low-quality visual inputs. This hinders the direct application of DP3 in our experimental setting. To ensure a fair comparison, we evaluate both \ourshort and DP3 on two short-horizon real-world tasks both using the learning-from-scratch~(LFS) encoder. The results are summarized in Table~\ref{table: h3dp_dp3}.

It is evident that DP3 underperforms compared to \ourshort in both tasks, highlighting \ourshort's ability to robustly extract meaningful features from RGB-D inputs, even when the quality of visual input is suboptimal. Furthermore, we empirically find that employing spatially sparse convolution provides better performance than the DP3-style encoder, suggesting a promising direction for improving point cloud encoding in low-fidelity settings.

\subsection{Inference Speed}

\begin{wraptable}{r}{0.55\textwidth}
    \renewcommand\tabcolsep{8.0pt}
    \centering
    \caption{\textbf{Comparison of inference speeds for DP, DP3 and \bfourshort in simulation tasks.} The result indicates that additional operations introduced in \ourshort are lightweight compared to the diffusion process.}
    \begin{tabular}{lccc}
        \toprule
        {Method} & DP & DP3 & \ourshort \\
        \midrule
        Inference Speed (FPS) & 11.1 & 12.2 & 12.0 \\
        \bottomrule
    \end{tabular}
    \label{table:infer}
\end{wraptable}

As shown in Table~\ref{table:infer}, we evaluate the inference speed of different methods within simulated environments. The results indicate that the primary bottleneck of the inference speed of \ourshort lies in the diffusion process itself, whereas the additional operations introduced for processing visual inputs and managing multi-scale representations incur only minimal computational overhead. A corresponding analysis of inference speed in real-world scenarios is available in Appendix~\ref{app:higherinference}.

\renewcommand{\arraystretch}{1.2} 
\begin{table}[ht] \caption{\textbf{Success rates on 44 simulation tasks}. Results of four different methods for each task are provided in this table. The summary across domains is shown in Table \ref{table:sim-results}.}
\label{table:sim_results_total}
\centering
\renewcommand\tabcolsep{5.0pt}
\resizebox{1.0\textwidth}{!}{
\begin{tabular}{c|cccccccc}
\toprule[0.5mm]
\begin{tabular}[c]{@{}c@{}}\multicolumn{1}{c}{\multirow{2}{*}{Method  $\backslash$ Tasks}}\end{tabular} & \multicolumn{8}{c}{\textbf{MetaWorld~\cite{yu2020meta}~(Medium)}} \\
& \small Basketball & \small Bin Picking & \small Box Close &\small Coffee Pull &\small Coffee Push & \small Hammer & \small Soccer &\small  Push Wall \\
\midrule 
\begin{tabular}[c]{@{}c@{}} \bfourshort  \end{tabular} &  \best{100} & \best{100} & \best{98} & \best{100} & \best{100} & \best{100} & \best{85} & \best{100} \\
\begin{tabular}[c]{@{}c@{}} DP \end{tabular}  & \best{100} & 96 & 83 & 82 & 84 & 64 & 43 & 76\\
\begin{tabular}[c]{@{}c@{}}DP~(w/ depth)\end{tabular} & \best{100} & 98 & 77 & 79 & 79 & 64 & 37 & 70\\
\begin{tabular}[c]{@{}c@{}} DP3 \end{tabular} &  \best{100} & \best{100} & 78 & \best{100} & \best{100} & 97 & 57 & 95\\
\midrule[0.3mm]
\end{tabular}
}
\resizebox{1.0\textwidth}{!}{
\begin{tabular}{c|ccc|ccccc}
\begin{tabular}[c]{@{}c@{}}\multicolumn{1}{c}{\multirow{2}{*}{Method  $\backslash$ Tasks}}\end{tabular} & \multicolumn{3}{c|}{\textbf{MetaWorld~(Medium)}} & \multicolumn{5}{c}{\textbf{MetaWorld~(Hard)}}\\
& \small Peg Insert Side  &\small Sweep &\small Sweep Into &\small Assembly &\small Hand Insert &\small Pick Out of Hole &\small Pick Place &\small Push\\
\midrule
\begin{tabular}[c]{@{}c@{}} \bfourshort  \end{tabular} & \best{98} & \best{100} & \best{100} & \best{100} & \best{100} & \best{40} & \best{99} & \best{100} \\
\begin{tabular}[c]{@{}c@{}} DP \end{tabular} & 62 & 96 & 74 & \best{100} & 73 & 13 & 0 & 77\\
\begin{tabular}[c]{@{}c@{}}DP~(w/ depth)\end{tabular} & 53 & 98 & \best{100} & \best{100} & 75 & 32 & 0 & 79\\
\begin{tabular}[c]{@{}c@{}} DP3 \end{tabular} & 92 & \best{100} & 61 & \best{100} & 37 & 30 & 0 & 96\\
\midrule[0.3mm]
\end{tabular}
}
\resizebox{1.0\textwidth}{!}{
\begin{tabular}{c|ccccc|cccc}
\begin{tabular}[c]{@{}c@{}} \multicolumn{1}{c}{\multirow{2}{*}{Method  $\backslash$ Tasks}} \end{tabular} & \multicolumn{5}{c|}{\textbf{MetaWorld~(Hard++)}} & \multicolumn{4}{c}{\textbf{DexArt~\cite{bao2023dexart}}} \\
&\small Shelf Place &\small Diassemble &\small Stick Pull &\small Stick Push &\small Pick Place Wall &\small Laptop &\small Faucet &\small Toilet &\small Bucket \\
\midrule
\begin{tabular}[c]{@{}c@{}} \bfourshort\end{tabular} & \best{100} & 96 & \best{83} & \best{100} & \best{100}  & \best{81} & \best{34} & 70 & \best{28} \\
\begin{tabular}[c]{@{}c@{}} DP \end{tabular}  & 20 & 81 & 64 & 70 & 55 & 69 & 23 & 58 & 27 \\
\begin{tabular}[c]{@{}c@{}}DP~(w/ depth)\end{tabular} & 29 & 76 & 71 & \best{100} & 80 & 63 & 20 & 62 & 23\\
\begin{tabular}[c]{@{}c@{}} DP3 \end{tabular} & 86 & \best{98} & 61 & \best{100} & 97 & 80 & 33 & \best{79} & 27\\
\midrule[0.3mm]
\end{tabular}
}
\resizebox{1.0\textwidth}{!}{
\begin{tabular}{c|ccc|cccc}
\begin{tabular}[c]{@{}c@{}} \multicolumn{1}{c}{\multirow{2}{*}{Method  $\backslash$ Tasks}} \end{tabular} & \multicolumn{3}{c|}{\textbf{Adroit~\cite{rajeswaran2017learning}}} & \multicolumn{4}{c}{\textbf{ManiSkill~\cite{gu2023maniskill2}~(Rigid)}}\\
&\small Hammer &\small Door &\small Pen  &\small  Peg Insertion Side (Grasp) &\small Peg Insertion Side (Align) &\small Pick Cube &\small Turn Faucet \\
\midrule
\begin{tabular}[c]{@{}c@{}} \bfourshort  \end{tabular} & \best{100} & \best{79} & \best{83} & 88 & \best{15} & \best{85} & \best{73} \\
\begin{tabular}[c]{@{}c@{}} DP \end{tabular} & 95 & 69 & 73 & 78 & 7 & 17 & 8  \\
\begin{tabular}[c]{@{}c@{}}DP~(w/ depth)\end{tabular} & \best{100} & 66 & 62 & \best{93} & 12 & 33 & 23 \\
\begin{tabular}[c]{@{}c@{}} DP3 \end{tabular} & \best{100} & 71 & 81 & 63 & 12 & 10 & 48 \\
\midrule[0.3mm]
\end{tabular}
}
\resizebox{1.0\textwidth}{!}{
\begin{tabular}{c|cccc|ccc}
\begin{tabular}[c]{@{}c@{}}\multicolumn{1}{c}{\multirow{2}{*}{Method  $\backslash$ Tasks}}\end{tabular} & \multicolumn{4}{c|}{\textbf{ManiSkill~(Deformable)}} & \multicolumn{3}{c}{\textbf{RoboTwin~\cite{mu2024robotwin}}}\\
& \small  Excavate &\small Hang &\small Pour &\small Fill & \small Apple Cabinet Storage &\small Dual Bottles Pick (Easy) & \small Dual Bottles Pick (Hard)  \\
\midrule
\begin{tabular}[c]{@{}c@{}} \bfourshort  \end{tabular} & \best{38} & \best{93} & \best{8} & \best{98} & \best{98} & 48 & \best{53} \\
\begin{tabular}[c]{@{}c@{}} DP \end{tabular} & 2 & 52 & 0 & 36 & 73 & 53 & 28 \\
\begin{tabular}[c]{@{}c@{}}DP~(w/ depth)\end{tabular} & 23 & 78 & 7 & 72 & 2 & 33 & 25\\
\begin{tabular}[c]{@{}c@{}} DP3 \end{tabular} & 15 & 80 & 0 & 12 & 55 & \best{55} & 42 \\
\midrule[0.3mm]
\end{tabular}
}
\resizebox{1.0\textwidth}{!}{
\begin{tabular}{c|ccccc|c}
\begin{tabular}[c]{@{}c@{}} \multicolumn{1}{c}{\multirow{2}{*}{Method  $\backslash$ Tasks}} \end{tabular} & \multicolumn{5}{c|}{\textbf{RoboTwin}} & \multicolumn{1}{c}{\multirow{2}{*}{\textbf{Average}}} \\
& \small Block Handover & \small Block Hammer Beat &  \small Diverse Bottles Pick  & \small Pick Apple Messy & \small Tool Adjust   \\
\midrule
\begin{tabular}[c]{@{}c@{}} \bfourshort  \end{tabular} & 70 & \best{85} & 25 & \best{35} & \best{45} & \textcolor{ourred2}{\ddbf{75.6}{18.6}}\\
\begin{tabular}[c]{@{}c@{}} DP \end{tabular} & 28 & 0 & 0 & 0 & 0 & \dd{48.1}{23.1}  \\
\begin{tabular}[c]{@{}c@{}}DP~(w/ depth)\end{tabular} & 0 & 0 & 2 & 7 & 32 & \dd{52.8}{22.2} \\
\begin{tabular}[c]{@{}c@{}} DP3 \end{tabular} & \best{85} & 47 & \best{30} & 8 & \best{45} & \dd{59.3}{24.9} \\
\bottomrule[0.5mm]
\end{tabular}
}
\end{table}

\renewcommand{\arraystretch}{1.2} 
\begin{table}[ht] \caption{\textbf{Success rates of experts on 44 simulation tasks.} We evaluate 200 episodes for each task. For ManiSkill tasks, the demonstrations are provided officially, and we record the success rates as 100\%. The final average result is obtained by averaging across all benchmarks.}
\label{table:expert}
\centering
\renewcommand\tabcolsep{5.0pt}
\resizebox{1.0\textwidth}{!}{
\begin{tabular}{c|cccccccc}
\toprule[0.5mm]
\begin{tabular}[c]{@{}c@{}}\multicolumn{1}{c}{\multirow{2}{*}{Method  $\backslash$ Tasks}}\end{tabular} & \multicolumn{8}{c}{\textbf{MetaWorld~\cite{yu2020meta}~(Medium)}} \\
& \small Basketball & \small Bin Picking & \small Box Close &\small Coffee Pull &\small Coffee Push & \small Hammer & \small Soccer &\small  Push Wall \\
\midrule 
\begin{tabular}[c]{@{}c@{}} Expert  \end{tabular} & 100.0 & 97.0 & 90.0 & 100.0 & 100.0 & 100.0 & 90.5 & 100.0 \\
\midrule[0.3mm]
\end{tabular}
}
\resizebox{1.0\textwidth}{!}{
\begin{tabular}{c|ccc|ccccc}
\begin{tabular}[c]{@{}c@{}}\multicolumn{1}{c}{\multirow{2}{*}{Method  $\backslash$ Tasks}}\end{tabular} & \multicolumn{3}{c|}{\textbf{MetaWorld~(Medium)}} & \multicolumn{5}{c}{\textbf{MetaWorld~(Hard)}}\\
& \small Peg Insert Side  &\small Sweep &\small Sweep Into &\small Assembly &\small Hand Insert &\small Pick Out of Hole &\small Pick Place &\small Push\\
\midrule
\begin{tabular}[c]{@{}c@{}} Expert  \end{tabular} & 92.0 & 100.0 & 90.0 & 100.0 & 100.0 & 100.0 & 100.0 & 100.0\\
\midrule[0.3mm]
\end{tabular}
}
\resizebox{1.0\textwidth}{!}{
\begin{tabular}{c|ccccc|cccc}
\begin{tabular}[c]{@{}c@{}} \multicolumn{1}{c}{\multirow{2}{*}{Method  $\backslash$ Tasks}} \end{tabular} & \multicolumn{5}{c|}{\textbf{MetaWorld~(Hard++)}} & \multicolumn{4}{c}{\textbf{DexArt~\cite{bao2023dexart}}} \\
&\small Shelf Place &\small Diassemble &\small Stick Pull &\small Stick Push &\small Pick Place Wall &\small Laptop &\small Faucet &\small Toilet &\small Bucket \\
\midrule
\begin{tabular}[c]{@{}c@{}} Expert \end{tabular} & 99.5 & 92.5 & 95.0 & 100.0 & 99.5 & 86.5 & 58.0 & 66.5 & 80.0\\
\midrule[0.3mm]
\end{tabular}
}
\resizebox{1.0\textwidth}{!}{
\begin{tabular}{c|ccc|cccc}
\begin{tabular}[c]{@{}c@{}} \multicolumn{1}{c}{\multirow{2}{*}{Method  $\backslash$ Tasks}} \end{tabular} & \multicolumn{3}{c|}{\textbf{Adroit~\cite{rajeswaran2017learning}}} & \multicolumn{4}{c}{\textbf{ManiSkill~\cite{gu2023maniskill2}~(Rigid)}}\\
&\small Hammer &\small Door &\small Pen  &\small  Peg Insertion Side (Grasp) &\small Peg Insertion Side (Align) &\small Pick Cube &\small Turn Faucet \\
\midrule
\begin{tabular}[c]{@{}c@{}} Expert  \end{tabular} & 99.0 & 100.0 & 97.0 & 100.0 & 100.0 & 100.0 & 100.0\\
\midrule[0.3mm]
\end{tabular}
}
\resizebox{1.0\textwidth}{!}{
\begin{tabular}{c|cccc|ccc}
\begin{tabular}[c]{@{}c@{}}\multicolumn{1}{c}{\multirow{2}{*}{Method  $\backslash$ Tasks}}\end{tabular} & \multicolumn{4}{c|}{\textbf{ManiSkill~(Deformable)}} & \multicolumn{3}{c}{\textbf{RoboTwin~\cite{mu2024robotwin}}}\\
& \small  Excavate &\small Hang &\small Pour &\small Fill & \small Apple Cabinet Storage &\small Dual Bottles Pick (Easy) & \small Dual Bottles Pick (Hard)  \\
\midrule
\begin{tabular}[c]{@{}c@{}} Expert  \end{tabular} & 100.0 & 100.0 & 100.0 & 100.0 & 96.0 & 97.0 & 55.5\\
\midrule[0.3mm]
\end{tabular}
}
\resizebox{1.0\textwidth}{!}{
\begin{tabular}{c|ccccc|c}
\begin{tabular}[c]{@{}c@{}} \multicolumn{1}{c}{\multirow{2}{*}{Method  $\backslash$ Tasks}} \end{tabular} & \multicolumn{5}{c|}{\textbf{RoboTwin}} & \multicolumn{1}{c}{\multirow{2}{*}{\textbf{Average}}} \\
& \small Block Handover & \small Block Hammer Beat &  \small Diverse Bottles Pick  & \small Pick Apple Messy & \small Tool Adjust   \\
\midrule
\begin{tabular}[c]{@{}c@{}} Expert  \end{tabular} & 98.0 & 97.0 & 72.0 & 88.5 & 86.5 & 93.9\\
\bottomrule[0.5mm]
\end{tabular}
}
\end{table}

\end{document}